\documentclass{article} 
\usepackage{times}

\usepackage{amsmath,amsfonts,bm}

\def\eqref#1{equation~\ref{#1}}

\def\1{\bm{1}}

\DeclareMathAlphabet{\mathsfit}{\encodingdefault}{\sfdefault}{m}{sl}
\SetMathAlphabet{\mathsfit}{bold}{\encodingdefault}{\sfdefault}{bx}{n}

\usepackage{subcaption}
\usepackage{hyperref}
\usepackage{url}
\usepackage{makecell}
\usepackage{graphicx}
\usepackage{multirow}
\usepackage{xspace}
\usepackage{natbib}

\newtheorem{example}{Template}[section]

\title{\card: Evaluating Causal Reasoning Capabilities of Large Language Models}

\author{
    Ruibo Tu\footnote{The author contributed to this work when he was with at KTH Royal Institute of Technology. He is currently with Qlik.} , Hedvig Kjellstr\"om, Gustav Eje Henter \\
    KTH Royal Institute of Technology, Stockholm, Sweden \\
    \texttt{\{ruibo, hedvig, ghe\}@kth.se} \\
    \and
    Cheng Zhang \footnote{The author contributed to this work mainly when she was with Microsoft Research. She is currently with Meta.} \\
    Microsoft Research, Cambridge, UK \\
    \texttt{cheng1001cheng@gmail.com}
}

\newcommand{\mistral}{Mistral-7B\xspace} 
\newcommand{\llama}{Llama3-8B\xspace} 
\newcommand{\gemma}{Gemma2-9B\xspace} 
\newcommand{\qwen}{Qwen2-7B\xspace} 
\newcommand{\mixtral}{Mixtral-8$\times$7B\xspace} \newcommand{\card}{\texttt{CARL-GT}\xspace}

\begin{document}

\maketitle

\begin{abstract}
Causal reasoning capabilities are essential for large language models (LLMs) in a wide range of applications, such as education and healthcare. But there is still a lack of benchmarks for a better understanding of such capabilities. Current LLM benchmarks are mainly based on conversational tasks, academic math tests, and coding tests. Such benchmarks evaluate LLMs in well-regularized settings, but they are limited in assessing the skills and abilities to solve real-world problems. 
In this work, we provide a benchmark, named by \card, which evaluates \underline{CA}usal \underline{R}easoning capabilities of large \underline{L}anguage models using \underline{G}raphs and \underline{T}abular data \footnote{Our benchmark is available at URL \url{https://github.com/TURuibo/CauTabBench/tree/main/carl_gt}.}. 
The benchmark has a diverse range of tasks for evaluating LLMs from causal graph reasoning, knowledge discovery, and decision-making aspects. 
In addition, effective zero-shot learning prompts are developed for the tasks. 
In our experiments, we leverage the benchmark for evaluating open-source LLMs and provide a detailed comparison of LLMs for causal reasoning abilities. We found that LLMs are still weak in casual reasoning, especially with tabular data to discover new insights.
Furthermore, we investigate and discuss the relationships of different benchmark tasks by analyzing the performance of LLMs. The experimental results show that LLMs have different strength over different tasks and that their performance on tasks in different categories, i.e., causal graph reasoning, knowledge discovery, and decision-making, shows stronger correlation than tasks in the same category.

\end{abstract}

\section{Introduction}
Causal reasoning ability is essential for large language models (LLMs). Firstly, it is a necessary property for reasoning with an action/intervention and is crucial for decision-making applications, ranging from education and healthcare to embodied AI systems, such as robotics and self-driving cars \citep{scholkopf2021toward,gupta2024essential}. 
Secondly, causal reasoning is highly related to mathematical reasoning. It often requires advanced mathematical reasoning abilities and the processing of complex data, such as tabular datasets and graphs. Traditionally, applying data-driven algorithms, such as causal discovery and inference methods, to structured data has been extensively used to augment human expert skills \citep{peters2017elements}.
Ultimately, if LLMs can effectively reason about both factual and counterfactual scenarios while understanding causal relationships, they can mitigate hallucinations, enhance interpretability, and significantly improve their decision-making capabilities, especially in agent-based applications.

Nevertheless, the current evaluation of LLMs primarily focuses on general reasoning tasks, including conversational problem-solving and challenges inspired by academic math and coding tests \citep{thawani2021representing,lu2024mathvista}. 
However, there is a notable lack of benchmarking tasks rigorously assessing causal reasoning capabilities. 
The absence of such benchmarks leaves a critical gap in understanding the limitations of LLMs and improving LLMs on causal reasoning.
Therefore, we introducing a benchmark explicitly targeting causal reasoning that would not only advance the evaluation of LLM reasoning capabilities but also help mitigate risks associated with their deployment in real-world applications.

\begin{figure}[t]
    \centering
    \begin{subfigure}{0.34\textwidth}
    \includegraphics[width=\linewidth]{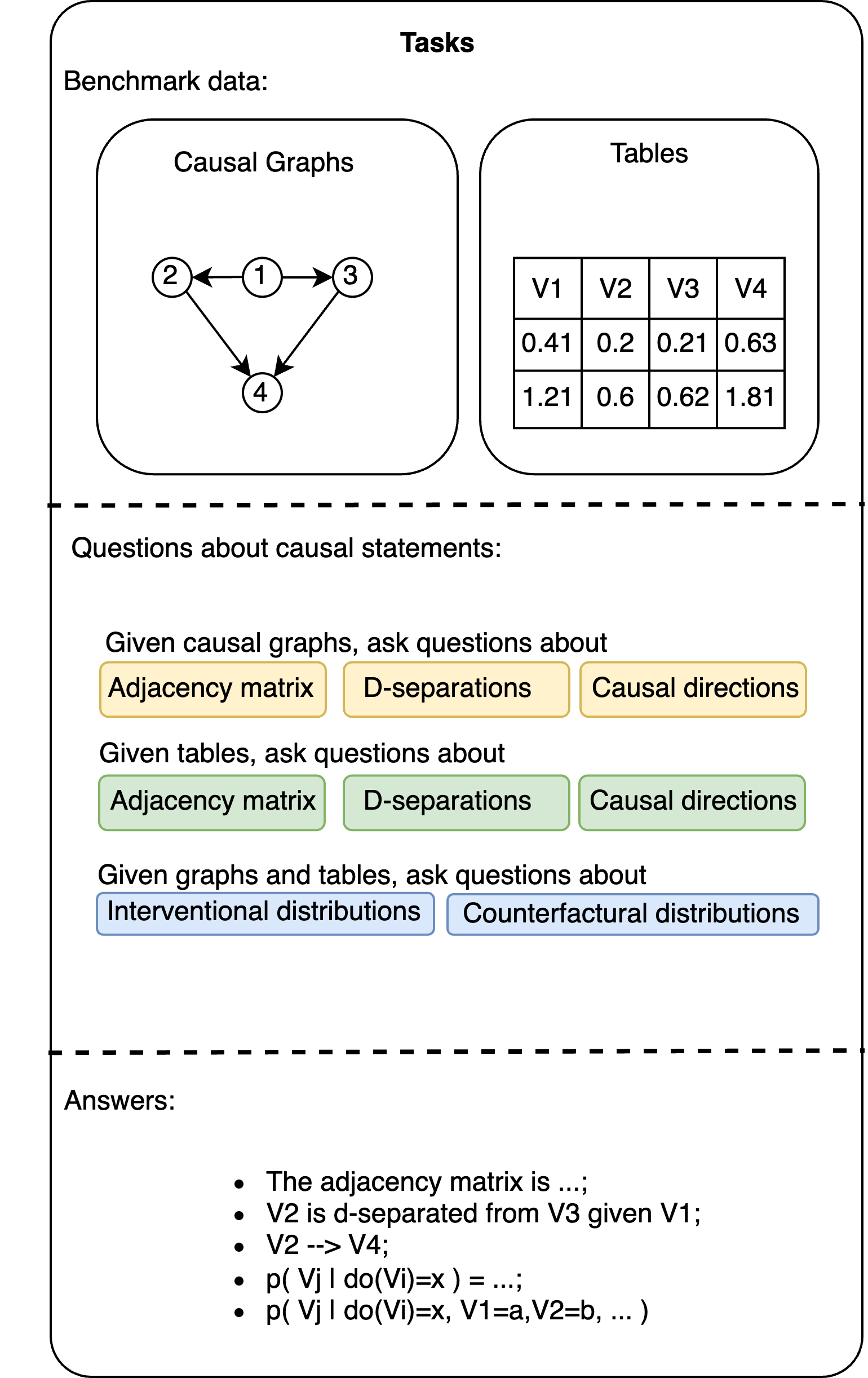}
    \caption{Benchmark data and parsers to compare with ground-truth answers.
    }
    \label{fig:composition}
\end{subfigure}
\hfill
\begin{subfigure}{0.63\textwidth}
    \includegraphics[width=\textwidth]{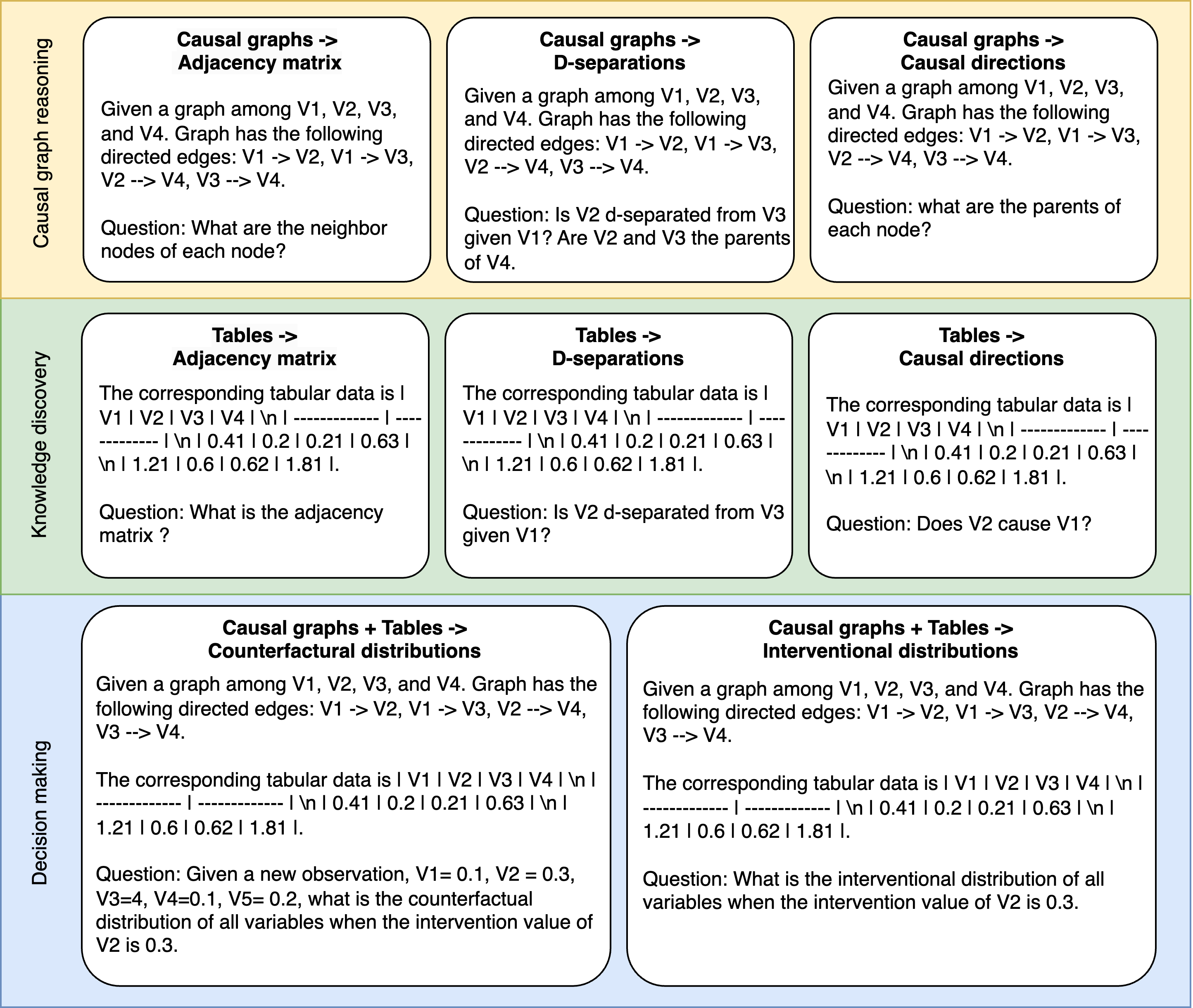}
    \caption{Prompt examples of our benchmark tasks for measuring the capabilities for causal graph reasoning, knowledge discovery, and decision making.}
    \label{fig:all_tasks}
\end{subfigure}
    \caption{Our benchmark, \card, for evaluating causal reasoning capabilities of LLMs from three aspects, i.e., causal graph reasoning, knowledge discovery, and decision making. 
    Fig. \ref{fig:composition} shows that a task consists of input data, such as causal graphs and tables, questions about causal statements, ground-truth answers, and parsers for evaluation.
    Furthermore, Fig. \ref{fig:all_tasks} shows that each aspect has multiple tasks, e.g., the causal statements about adjacency matrix, d-separation, and causal direction, are used for causal graph reasoning and knowledge discovery evaluation and the statements about intervention and counterfactual reasoning are used for decision-making evaluation.}
    \label{fig:benchmark_tasks}
\end{figure}

More specifically, to measure the LLM capabilities for causal reasoning, we consider three categories of causal-aware tasks.
First, \emph{Causal Graph Reasoning} tasks are designed to test whether LLMs can perform basic causal graph reasoning when provided with predefined causal graphs.
Second, \emph{Knowledge Discovery} focuses on evaluating the ability of LLMs to infer causal relationships from observational tabular data. 
Finally, \emph{Decision Making} assesses whether LLMs can produce accurate outputs when subjected to variable interventions, corresponding to intervention and counterfactual reasoning in the three levels of the reasoning ladder \citep{pearl2016causal}, applicable to treatment effect estimation and decision-making. In this work, our contributions are that
\begin{itemize}
    \item We propose a benchmark, \card, formatting causal graph data and numerical tabular data as texts for LLM evaluations as shown in Fig. \ref{fig:benchmark_tasks}. \card includes different categories of causal skills, causal graph reasoning, knowledge discovery, and decision-making, covering a range of scenarios where causal reasoning skills are essential, such as scientific discovery or healthcare decision-making, and ensuring a diverse set of tasks that reflect the general causal reasoning ability of LLMs. 
    \item We evaluate LLMs with \card by developing effective prompts in a zero-shot manner. The experimental results show that LLMs have different strength over different tasks. Roughly speaking, \mixtral \citep{jiang2023mistral} is better at causal graph reasoning while \mistral \citep{jiang2023mistral} is better at the other tasks than \qwen \citep{qwen} and \gemma  \citep{gemma_2024}. Moreover, we investigate relationships among different tasks by studying the relationships of the performance of LLMs on the tasks. We find that tasks about the same topics in different categories have stronger connection with each other than the tasks in the same categories.
\end{itemize}

\section{Related Work} \label{sec:related_works}
Multiple leader-boards including standard LLM benchmarks have been used for evaluating LLM models, such as LMSYS Chatbot Arena \citep{chiang2024chatbot} and the open LLM leaderboard \citep{open-llm-leaderboard-v2} on Huggingface. The drawback of focusing on leader-boards is seen by the community according to Goodhart's Law which indicates ``when a measure becomes a target, it ceases to be a good measure." Consequently, more and more benchmarks are proposed with more extensive evaluation on different aspects, such as general reasoning \citep{clark2018think,rein2023gpqa}, math reasoning \citep{cobbe2021training,hendrycks2021measuring}, multi-modal reasoning \citep{lu2023mathvista}, long context \citep{lee2024can}, and function calling \citep{berkeley-function-calling-leaderboard}. In our work, we complement these new evaluations regarding causal reasoning ability, which is an essential ability for human-level intelligence and requires many sub-skills including multi-step reasoning, mathematically reasoning, graph reasoning, tabular data processing etc. We will first introduce the related benchmarks on mathematical reasoning, numerical reasoning, graph reasoning, and causal reasoning in this section.

\paragraph{Benchmarks on mathematical and numerical reasoning.}
\citet{lu2024mathvista} have recently proposed benchmark tasks for evaluating the LLMs' capabilities for algebraic reasoning, arithmetic reasoning, geometry reasoning, logical reasoning, numeric common sense, scientific reasoning, and statistical reasoning. \citet{thawani2021representing} evaluate the numerical reasoning abilities of LLMs. There are other similar benchmark datasets based on math tests \citep{lu2021inter,hendrycks2021measuring,lindstrom2022clevr}. Furthermore, some works step further and consider more challenging scenarios based on real-world applications, such as numerical predictive tasks including density estimation, probabilistic multivariate regression \citep{requeima2024llm}, and time-series forecasting \citep{gruver2024large}. Moreover, some LLM benchmarks evaluate on the understanding, fact verification, and summerization of (semi-)structured data including both numerical values and texts \citep{pasupat-liang-2015-compositional,iyyer-etal-2017-search,parikh-etal-2020-totto,FEVEROUS21,FeTaQA}. Our benchmark is inspired by real-world applications and evaluates the causal reasoning abilities given observational tabular data and/or causal graph data.

\paragraph{Benchmarks on graph reasoning.} A few works propose benchmarks for benchmarking LLMs on reasoning graph data. \citet{wang2024can} propose a benchmark for evaluating the capability of LLMs with the text descriptions of general graph data. \citet{fatemi2024talk} evaluate LLMs with simple graph tasks at node, edge, and graph levels, such as node counting, edge existence, and cycle check. Besides general graphs, it also considers graphs in different context, such as social network. Similarly, \citet{Jiayan2024GPT4Graph} evaluate LLMs with simple general graph tasks and semantic tasks, such as node classification and graph classification of knowledge graphs. \citet{changbo23} also benchmark LLMs on general graph tasks about connectivity, neighbor classification, node degree, pattern (wedge or a triangle) matching, and shortest paths. Moreover, similar tasks are also used in a multi-modality setting for evaluating multi-modal LLMs. \citet{wei2024rendering,wei2024gita,das-etal-2024-modality} encode graph data as images or other modalities together with texts. In contrast, we focus on causal graphs, and propose benchmark tasks in the context of causality for better understanding the capability of LLMs on causal reasoning. 

\paragraph{Benchmarks on causal reasoning.} \citet{jin2023cladder, ze2023causal, Jinetal24, caueva24} propose benchmarks for measuring causal reasoning capabilities based on text descriptions of causal statements and causal graphs. \citet{jin2023cladder} evaluate the performance of LLMs with causal-inference question-answering. It transforms causal inference problems (e.g., intervention and counterfactual inference) in natural language, provides essential computation quantities as stories, and requires LLMs to answering such causal inference-related questions. \citet{Jinetal24} evaluates the performance of LLMs on distinguishing causal relationships from correlations. It generates causal graphs and derives correlation and d-seperation relationships that are further used for generating correlation and causation statements. And LLMs are required to distinguishing causal statements from correlation statements. Similarly, \citet{CLEAR24} verbalize causal graphs and causal relationships for evaluating LLMs on causal graph understanding. The tasks are about association, causal discovery, intervention reasoning, and counterfactual reasoning. Furthermore, with similar tasks, \citet{CELLO24} propose a multi-modal benchmark including scene images and scene graphs. Nevertheless, our benchmark further considers a more realistic and challenging setting for measuring causal reasoning abilities that asks LLMs to process observational tabular data and/or causal graph data and answer questions from different perspectives of causal reasoning and decision making.

\section{Benchmark \card}
We propose a comprehensive evaluation benchmark for causal reasoning on numerical tabular data, graphs, and texts. We consider the causal reasoning capabilities from three types: 
1) \textit{Causal graph reasoning} that understands the properties and relationships of causal graphs; 
2) \textit{Knowledge discovery} that identifies causal graphs from observational tabular data; 
3) \textit{Decision making} that requires to estimating functional relationships among concerned variables and estimating intervention distributions and counterfactual reasoning for causal inference. 
Furthermore, guided by the reasoning types, the benchmark tasks are
1) causal graph question-answering given causal graphs for the evaluation of causal graph reasoning; 
2) d-separation estimation and causal direction estimation given observational data for the evaluation of causal knowledge discovery;
3) causal inference, i.e., intervention and counterfactual inference, given either observational data or both observational data and causal graphs for the evaluation of decision-making.

\subsection{Benchmark datasets}
Our benchmark examines LLMs by requiring them determining causal statements and computing causal quantities based on given benchmark data. 
As shown in Fig. \ref{fig:composition}, benchmark data consist of causal graph data and corresponding observational data in the form of  tables. 
The causal graphs used in our benchmark data are directed acyclic graphs of which the nodes represent the variables of interests and the edges represent the causal relationships between variables, 
denoted by $\mathcal{G}=\{G_g = (V_g,E_g)\}_{g=1}^{N_{\texttt{g}}}$ where $N_{\texttt{g}}$ is the total number of causal graphs, $V_g$ is the node sets, and $E_g$ is the edge set of graph $G_g$. 
For example, node 1 is the parent of node 2, which represents that node 1 is the direct cause of node 2. 
The tables are corresponding to causal graphs and generated with structural causal models defined according to causal graphs.
Each column of a table is corresponding to a node of a causal graph and each row of a table is corresponding to an instance following the data generation process defined with a causal graph. 
The tables are denoted by $\mathcal{D}_{G_g}=\{(x_1,x_2,...,x_{N_\texttt{V}})_i\}_{i=1}^{N_{\texttt{sz}}}$ where $N_\texttt{V}$ represents the number of columns and $N_{\texttt{sz}}$ represents the number of rows.

Besides benchmark data, our benchmark also includes causal quantities and causal statements. They are used in benchmark tasks and can reflect different perspectives of causal reasoning capabilities. The causal statements are about adjacency matrices, d-separations, and causal directions that have an increasing of difficulties to be identified from observational data. Estimating adjacency matrices requires correctly estimating whether each pair of variables is d-separated or not, while estimating d-separations requires further knowing conditioning on which variable set. And estimating causal directions requires more information about data generation processes than estimating d-separation relationships. Moreover, the causal quantities about intervention and counterfactual reasoning are subject to interventions. Such quantities are the foundation for causal inference applications, such as average treatment effect, natural direct effect, and natural indirect effect.

\subsection{Benchmark tasks}
Our benchmark tasks are composed by three parts, prompted benchmark data, examined causal-related concepts, and ground-truth answers as shown in Fig. \ref{fig:composition}. Based on different prompted data and causal-related concepts, benchmark tasks can be used for evaluating the LLMs' capabilities for causal graph reasoning, causal knowledge discovery, and decision-making summerized in Fig. \ref{fig:all_tasks}.

\paragraph{Causal graph reasoning.}
Our benchmark tasks for evaluating causal graph reasoning ability of LLMs are similar as the benchmarks \citep{jin2023cladder,CLEAR24} introduced in Sec. \ref{sec:related_works}. 
The tasks are designed coherently with the other two types of tasks. And based on the current evaluation of causal knowledge discovery and decision making, the minimum necessary concepts for causal graph reasoning are adjacency matrices, d-separation relationships, and causal directions. Correctly understanding and estimating such relationships is the key for causal knowledge discovery and decision making. 
For example, d-separation is the foundation for many advanced causal reasoning conditions, such as front-door and back-door criteria \citep{pearl2016causal}, and correctly estimating  d-separation relationships can help relax the influence of spurious correlation.
In the tasks, we prompt causal graphs $\mathcal{G}$ with different strategies, e.g., Fig. \ref{fig:all_tasks} demonstrates Strategy Expert \citep{fatemi2024talk}. We then ask LLMs about 
1) the adjacent nodes of each node; 
2) the d-separation relationships between pairs of nodes;
3) the parents of each node.
Furthermore, inspired by \citet{tu24}, the metrics are 
1) F1 score, precision, and recall for estimating adjacency matrices;
2) the AUC of ROC curves for estimating d-separations;
3) F1 score, precision, and recall for estimating causal graphs.

\paragraph{Knowledge discovery.}
The goal of causal discovery-based tasks is to examine whether LLMs can determine different causal-related relationships which are crucial for knowledge discovery from unseen datasets that are commonly tabular data. The causal-related relationships are adjacency matrices, d-separations, and causal directions. They have an increasing order of difficulties to be identified by causal discovery methods. In this way, the evaluation can tell that to which extend, LLMs can perform well on causal knowledge discovery. The numerical data $\mathcal{D}_{G_g}$ are serialized into different formats for LLMs. We consider MarkDown for the serialization. Fig. \ref{fig:all_tasks} demonstrates MarkDown format. The tasks are in a similar form as the causal reasoning tasks and only differ in the way that prompted data are numerical tables instead of causal graphs. The metrics are the same as the ones in causal reasoning tasks. Moreover, as for the task about causal direction of which the arrow heads are children (effects) and the arrow tails are parents (causes),  we use the accuracy for estimating causal directions.

\paragraph{Decision making.}
Causal-inference based tasks are used for measuring the decision-making performance. 
Predicting the consequences of actions is the key for effective and efficient data-driven decision-making systems \citep{gupta2024essential}. By understanding how interventions on certain variables have an influence on outcomes, LLMs can simulate and evaluate consequences of actions, linking causal reasoning to actionable insights. 
Similarly, counterfactual reasoning, which involves reasoning about "what-if" scenarios, helps assess the consequences of decisions that were not taken, improving the generalization and robustness of LLM agents \citep{richens2024robust}. Therefore, we propose the tasks based on intervention and counterfactual reasoning. Given numerical data $\mathcal{D}_{G_g}$ and corresponding causal graphs $\mathcal{G}_g$, we first impose interventions on some variables and LLMs are required to estimating the distributions of some other variables subject to the interventions. The ground-truth answers are computed by learning structural causal models. 
As for the counterfactual inference, we provide a sample as a given fact and then ask LLMs to estimate the counterfactual results subject to the counterfactual values. The metric is mean absolute errors between the ground-truth and the estimated expectation. The accurate estimation of intervention distributions and counterfactual reasoning indicates the successes in causal inference applications, such as average treatment effect estimation in healthcare.

\section{Experiments}
\subsection{Experimental setup}
\paragraph{Baseline LLMs.}
In all experiments, the baseline LLMs are text-only LLMs, 
\llama \citep{llama3modelcard},
\qwen \citep{qwen},
\gemma  \citep{gemma_2024},
\mistral \citep{jiang2023mistral}, and
\mixtral \citep{jiang2023mistral}. 
We benchmark LLMs to the proposed tasks in the zero-shot manner. 

\paragraph{Benchmark data configuration.}
We randomly generate direct acyclic graphs as causal graphs, and then generate tabular data according to the causal graphs following \citep{tu24}. More specifically, the number of nodes of causal graphs are $10$ in Sec. \ref{sec:bench_results} and tabular data are generated by the linear uniform structural causal models with random parameters. The node number is $51$ for the experiments in Sec. \ref{sec:nodes}. The causal graphs are converted as texts with 
\begin{example}[A template format for causal graphs.]\label{promp:expert}
    A causal graph has nodes V0, V1, V2, V3, V4, V5, V6, V7, V8, and V9. And its edges are $V1 -> V2$, $V3 -> V4$, ..., $V8 -> V9$.
\end{example}
The tables are prompted in the Markdown format and $50$ rows are included in prompts. Each LLM is required to answer questions of benchmark tasks. The prompts depend only on causal graphs in causal graph reasoning tasks, only on tables in causal knowledge discovery tasks, and on both tables and causal graphs in causal inference tasks. 

\paragraph{Evaluation procedures.}
We take two steps to extract final answers from LLM responses for the evaluation. We first ask the baseline methods to reply to the questions of benchmark tasks and then ask them to extract the answers in specific formats for computing the metric values. Since LLMs can directly get the answers to the causal knowledge discovery questions, we apply the two-step manner to the other tasks. Prompt \ref{promp:questions} is the prompt questions and Prompt \ref{promp:refinement} is used for extracting the final answers based on LLM responses. 
Especially, in Prompt \ref{promp:questions}, we use the Markdown format for the \texttt{\{tabular\_data\}} and the ``Expert" format \citep{fatemi2024talk} for \texttt{\{graph\_data\}}, e.g., ``You are a graph analyst and you have been given a graph G among A, B, C, D, E, F, G, H, and I. G has the following edges: $A -> B$, $A -> C$, ..., $H -> I$."

\begin{example}[Tasks questions.]\label{promp:questions} 
In the templates, \{graph\_data\} takes the input text of a causal graph in the benchmark datasets; \{tabular\_data\} takes a serialized table in the benchmark datasets. \{task question\} are the questions for different tasks. 

The template of causal graph reasoning questions is ``You are reasoning over causal graphs. \{graph\_data\}. \{task question\}." 

The template of causal discovery-based questions is ``You are reasoning over tables. Columns represent different nodes and rows represent different samples. The data are \{tabular\_data\}. \{task question\}. Answer the question with only yes or no." 

The template of causal inference-based questions is ``You are reasoning over tables and causal graphs and then answer causal inference questions. Columns represent different nodes and rows represent different samples. The data are \{tabular\_data\}. The causal graph is \{graph\_data\}. \{task question\}. You must provide a float number as the final results."

For example, questions for 
\begin{itemize}
    \item Estimating d-separations: Are V7 and V8 d-separated given V6?
    \item Estimating conditional independence: Are V5 and V9 conditionally independent given V4?
    \item Estimating causal directions: Between V3 and V5, is V3 the cause?
    \item Estimating counterfactual distributions:
    Given the values of V0 to V9 are 1.573, 1.116, 0.151, -0.063, -0.407, -0.443, 1.657, 1.585, 0.169, -0.275, what is the expectation of the distribution of V3 if the value of V0 is 1.122224260377778?
    \item Estimating intervention distributions: Given that the intervention value on V0 is 1.122224260377778, what is the expectation of the intervention distribution of V7? 
\end{itemize}
\end{example}

\begin{example}[Extracting answers from LLM responses.]\label{promp:refinement} 
As for causal graph reasoning tasks, the template of extracting adjacency matrices is that ``You are extracting neighbor information from given responses. The response is: \{response\}.  The result in the format: Neighbors of XXX are XXX, ... where XXX should be replaced by nodes"; for extracting d-separations is that ``The question is \{task question\}. The response is: \{response\}. Given the question and the response, summarize the answer as yes or no based on the response."

The template of extracting final answers of causal inference tasks 
is that ``You are extracting the final answer from a response of a model. The response is \{response\}. Summarize the answer as a float number from the response. The response is less than 10 tokens. The final answer is:".
\end{example}

\subsection{Experimental results}\label{sec:bench_results}
\paragraph{Causal graph reasoning.} 
The benchmarking results of causal graph reasoning tasks are shown in Tab. \ref{tab:causal_graph}. 
The adjacency matrices are recovered by specifying the neighbors of each node in the LLM responses. 
The d-separation estimation is evaluated in a binary classification manner and the d-separation and d-connection sets of questions are selected following \citep{tu24}. We create a balanced classification dataset with labels as either d-separated or d-connected and ask the baseline methods to determine whether provided relationships are d-separated or d-connected. 
The causal directions are recovered by specifying the parents of each node in the LLM responses.
As shown in Tab. \ref{tab:causal_graph}, the results of adjacency matrix estimation are much better than causal direction estimation and d-separation estimation of which a random classifier can have AUC result $0.5$. 
This is because the causal graphs are prompted in a way that the neighbors of each node are straightforward to be identified and LLMs only need to check edges one-by-one for getting the correct answers. In contrast, d-separation and d-connection relationships are much harder to be classified, which requires LLMs to reasoning based on multiple edge information and directions at the same time. Moreover, causal direction estimation further requires LLMs to distinguish the direction of edges, which makes the task more difficult than adjacency matrix estimation. By comparing different baseline LLMs, Qwen2 has the more satisfying results on both tasks than the other LLMs, while Mixtral 8x7B has the best performance of adjacency matrix estimation, Gemma2 9B has the best performance of d-separation estimation, and Llama3 has the best performance of causal direction estimation.
\begin{table}
    \centering     
    \caption{Benchmark on causal graph reasoning. ``Adj.", ``d-sep.", and ``dir." represent for adjacency matrix, d-separation, and causal direction estimation.
    As for adjacency matrix and causal direction estimation, LLMs are required to recover $10$ causal graphs. The metrics are F1 scores, recall, and precision. As for d-separation estimation, LLMs classify d-separated and d-connected relationships chosen from $10$ causal graphs. And the metric is the AUC of ROC curves.}\label{tab:causal_graph}
        \resizebox{0.99\columnwidth}{!}{
\begin{tabular}{c|ccc|c|ccc}
    \hline
    &\multicolumn{3}{c|}{Adj.}&D-sep.&\multicolumn{3}{c}{Dir.}\\\hline
	  	 & f1 $\uparrow$ 	 & precision $\uparrow$ 	& recall $\uparrow$  & AUC $\uparrow$ 	  	 & f1 $\uparrow$ 	 & precision $\uparrow$ 	& recall $\uparrow$\\ \hline
\llama &$0.75\pm0.14$ &$0.64\pm0.20$ &$ 0.96\pm0.05$&$0.475$&$\mathbf{0.48\pm0.15}$ &$0.33\pm0.13$ &$ 0.95\pm0.11$\\
\qwen &$0.90\pm0.05$ &$0.91\pm0.10$ &$ 0.90\pm0.05$&$0.514$ &$0.42\pm0.22$ &$0.33\pm0.21$ &$ 0.74\pm0.24$ \\
\mistral &$0.83\pm0.05$ &$0.87\pm0.09$ &$ 0.82\pm0.07$&$0.464$&$0.35\pm0.24$ &$0.25\pm0.17$ &$ 0.85\pm0.26$\\
\gemma &$0.81\pm0.16$ &$0.71\pm0.24$ &$1.00\pm0.00$&$\mathbf{0.562}$&$0.45\pm0.13$ &$0.30\pm0.11$ &$ 1.00\pm0.00$  \\
\mixtral  &$\mathbf{0.95\pm0.04}$ &$1.00\pm0.00$ &$ 0.92\pm0.07$&$0.482$&$0.30\pm0.20$ &$0.20\pm0.17$ &$ 1.00\pm0.00$\\
\hline
\end{tabular}}
\end{table}

\begin{table}
\centering     
\caption{Benchmark on knowledge discovery. As for d-separation and causal direction estimation, the answers are limited by yes or no. The questions are based on $10$ causal graphs. 
The metrics are the AUC of ROC curves for d-separation estimation and accuracy for causal direction estimation.
When there are unknown answers from LLMs, we replaced by random yes or no. Gemma only takes no more than 20 rows of tables as inputs, so we input tables with $20$ rows for Gemma while the others take input tables with $50$ rows. We do not include the results for Mixtral because of the limitation GPU memory.}\label{tab:table}
\begin{tabular}{c|cc}
\hline
	  	  & AUC $\uparrow$ 	 & Acc. $\uparrow$ 	 \\ \hline
\llama &$\mathbf{0.562}$ &$0.321$\\
\qwen  &$0.50$ &$0.00$ \\
\mistral  &$0.554$&$\mathbf{0.429}$\\
\gemma   &$0.50$ &$0.00$ \\
\mixtral &$-$&$0.00$\\
\hline
\end{tabular}
\end{table}

\paragraph{Knowledge discovery.}
We only include the experimental results of d-separation and causal direction estimation because zero-shot LLMs cannot work on adjacency matrix estimation by providing tables as input and asking ``what is the adjacency matrix?" or ``what are the neighbors of each nodes?" Because it takes multiple steps and it can be beneficial to use in-context learning or chain-of-thought for the task in the future work. The question prompts include tables with $50$ rows except the experiments of Gemma2 where the question prompts only have $20$ rows of tables. Because Gemma2 can only take at most $20$ rows of each table based on our hardware ($4$ \texttt{x Nvidia-A100} and $4$ \texttt{x Nvidia-A40}). Following \citep{tu24}, we select d-separation and d-connection sets as well as causal directions and then use AUC and accuracy as metrics. 
Different from the results in Tab. \ref{tab:causal_graph}, all methods at least have AUC value $0.5$ for d-separation estimation and \llama and \mistral are better than \qwen and \gemma. Moreover, the causal direction estimation turns out to be more challenging for LLMs, because \qwen, \gemma, and \mixtral cannot provide the answer to the questions and reply that the information is not enough to answer the questions. 
\begin{table}
    \centering     
    \caption{Benchmark on the interventional and counterfactual inference tasks. The answers with unknown responses are skipped. LLMs are prompted with $10$ causal graphs and corresponding tables with $50$ rows. 
    The metric is the mean absolute errors of predictions compared with the ground-truth results. ``inv-" and ``cf-" represent for intervention and counterfactual.
    We do not include the results for Gemma because of the limitation GPU memory.}\label{tab:ci}
    \begin{tabular}{c|cc}
    \hline
	  	 & inv-MAE $\downarrow$ 	  & cf-MAE $\downarrow$ 	 \\ \hline
\llama&$\mathbf{2.053 \pm 1.391}$&$1.586 \pm 1.060$\\
\qwen  &$2.612 \pm 1.011$ & $1.475 \pm 0.642$ \\
\mistral  &$2.145 \pm 1.437$&$1.500 \pm 0.661$\\
\gemma  &$-$ &$-$ \\
\mixtral &$2.409 \pm 1.407$&$\mathbf{1.125 \pm 0.359}$\\
\hline
\end{tabular}
   
\end{table}

\paragraph{Decision-making.} We evaluate the ability of decision making with causal inference tasks. To get the ground-truth for causal inference tasks, we estimate structural causal models with $15000$ benchmark data and then use the model to compute the intervention distributions subject to interventions and counterfactual results subject to counterfactual values. 
The expectation of ground-truth intervention distributions is estimated by averaging the sampled $1000$ data. 
Different from estimating intervention distributions, the ground-truth for counterfactual results is computed based on using invertible functional relationships such that they can be uniquely identified. We evaluate LLMs with $100$ intervention questions and $100$ counterfactual questions based on $10$ causal graphs and $10$ corresponding tables. The question prompts include $50$ rows of each table.
As shown in Tab. \ref{tab:ci}, \mistral has a satisfying overall results on both tasks while Mixtral has the best performance of counterfactual inference.

\subsection{Experiments with more nodes in causal graphs and more rows of tables}\label{sec:nodes}
We investigate the impact of number of nodes in causal graphs on the benchmark results and compare the performance of \qwen, \llama, and \mistral in Tab. \ref{tab:causal_graph_nodes}. 
We found that the benchmark tasks with more nodes in causal graphs are significantly more challenging for LLMs, and that their performance is in general much worse than the one with fewer nodes. As for Qwen, it has memory issues that input cannot take tables with $50$ rows; hence, the inputs take tables with $20$ rows in the experiments for knowledge discovery and decision making. Moreover, Llama cannot output answers for knowledge discovery results; hence, their values are $NA$ in the table. 
\begin{table}[!h]
    \centering     
    \caption{Comparing benchmark results based on causal graphs with $51$ and $10$ nodes. }\label{tab:causal_graph_nodes}
    \resizebox{0.99\columnwidth}{!}{
    \begin{tabular}{c|c|cc|cc|cc}
    \hline
 & &\multicolumn{2}{c|}{\mistral} &\multicolumn{2}{c|}{\llama} & \multicolumn{2}{c}{\qwen}\\\hline
 &Metrics &$51$ nodes & $10$ nodes &$51$ nodes&$10$ nodes   &  $51$ nodes &$10$ nodes\\\hline
 \multirow{7}{*}{Graph}&adj-f1 $\uparrow$ 	   &$0.34\pm0.11$ &$0.83\pm0.05$ &$0.46\pm0.06$&$0.75\pm0.14$   &$0.39\pm0.08$ &$0.90\pm0.05$ \\
 &adj-precision $\uparrow$ &$0.24\pm0.08$&$0.87\pm0.09$ &$0.31\pm0.05$&$0.64\pm0.20$  &$0.25\pm0.06$ &$0.91\pm0.10$ \\
 &adj-recall $\uparrow$     &$ 0.63\pm0.30$ &$ 0.82\pm0.07$&$ 0.98\pm0.02$&$ 0.96\pm0.05$&$ 0.91\pm0.05$&$ 0.90\pm0.05$\\ 
& dsep-AUC $\uparrow$    &  $0.493$&$0.464$& $0.509$&$0.475$& $0.583$&$0.514$   \\ 
&dir-f1 $\uparrow$ 	   &$0.52\pm0.18$ &$0.35\pm0.24$  &$0.46\pm0.22$ &$0.48\pm0.15$ &$0.09\pm0.06$ &$0.48\pm0.15$ \\
 &dir-precision $\uparrow$ 	    &$0.47\pm0.10$ &$0.25\pm0.17$  &$0.36\pm0.18$&$0.33\pm0.13$ &$0.11\pm0.10$  &$0.33\pm0.13$ \\
 &dir-recall $\uparrow$     &$ 0.62\pm0.23$ &$ 0.85\pm0.26$&$ 0.66\pm0.30$&$ 0.95\pm0.11$&$ 0.17\pm0.20$&$ 0.95\pm0.11$\\ \hline
\multirow{2}{*}{Table}&AUC $\uparrow$ &$0.497$ 	 &$0.554$&$NA$&$0.562$ &$0.5$ &$0.50$ \\
&Acc. $\uparrow$ &$0.1724$	 &$0.429$&$NA$&$0.321$&$0.0$ &$0.00$ \\\hline
\multirow{2}{*}{Table+Graph}&inv-MAE $\downarrow$&$1.645 \pm 0.832$&$2.145 \pm 1.437$&$9.071 \pm 8.793$ &$2.053 \pm 1.391$& $2.449 \pm 1.344$	 &$2.612 \pm 1.011$ \\ 
&cf-MAE $\downarrow$ & $5.003 \pm 2.102$&$1.500 \pm 0.661$& $8.572 \pm 6.770$&$1.586 \pm 1.060$&$0.971 \pm 0.393$   & $1.475 \pm 0.642$ \\\hline
\end{tabular}}
\end{table}

As shown in Tab. \ref{tab:num_rows}, we compare the performance of \qwen and \mistral with question prompts including $50$ and $100$ rows of tables. In general, there are no significant differences on the performance. This can be the reason that both situations with $50$ and $100$ rows are challenging such that LLMs cannot gain a significant improvement by getting $50$ more rows of tables in question prompts.

\begin{table}[h]
    \centering     
    \caption{The impact of the row number of tables in question prompts on the performance of knowledge discovery and decision making tasks. We compare the results of Qwen2 and Mistral with $50$ and $100$ rows serialized tables in the question prompts.}\label{tab:num_rows}
\begin{tabular}{c|cccc}
    \hline
	  	 & intv-MAE $\downarrow$	& cf-MAE $\downarrow$	 & AUC $\uparrow$	  & Acc. 	$\uparrow$ \\ \hline
\qwen-50-rows &$2.612 \pm 1.011$ & $\mathbf{1.475 \pm 0.642}$ & $0.50$  & $0$ \\
\qwen-100-rows  & $\mathbf{2.281 \pm 0.640}$ & $2.001 \pm 0.913$ & $0.50$  & $0$ \\\hline
\mistral-50-rows  &$2.145 \pm 1.437$&$\mathbf{1.500 \pm 0.661}$&$\mathbf{0.554}$&$0.429$\\
\mistral-100-rows  &$\mathbf{1.908 \pm 0.760}$&$2.732 \pm 2.228$&$0.542$&$\mathbf{0.449}$\\\hline
\end{tabular}
\end{table}

\subsection{Other factors for the performance on benchmark results}
Many factors can contribute to the less ideal results of LLMs. We applied LLMs to the benchmark tasks in a zero-shot learning manner without an extensive effort searching for the best prompts. For example, future works can try to use chain-of-thought with in-context learning examples for further improving the performance. Moreover, $50$ or $100$ rows of tables are included in the question prompts. Roughly speaking, Tab. \ref{tab:num_rows} shows that increasing the number of rows from $50$ to $100$ has no significant improvement on the performance. This may be due to the fact that both $50$ and $100$ rows provide insufficient observational data for both causal discovery and causal inference, which requires at least $500$ data for the more consistent results in practice. On the other hand, the number of rows of tables included in the question prompts is limited by maximum input tokens of LLMs and GPU memories.

\subsection{Relationships among benchmark tasks} 

\begin{figure}[h]
    \centering
    \begin{subfigure}{0.31\textwidth}
    \includegraphics[width=\linewidth]{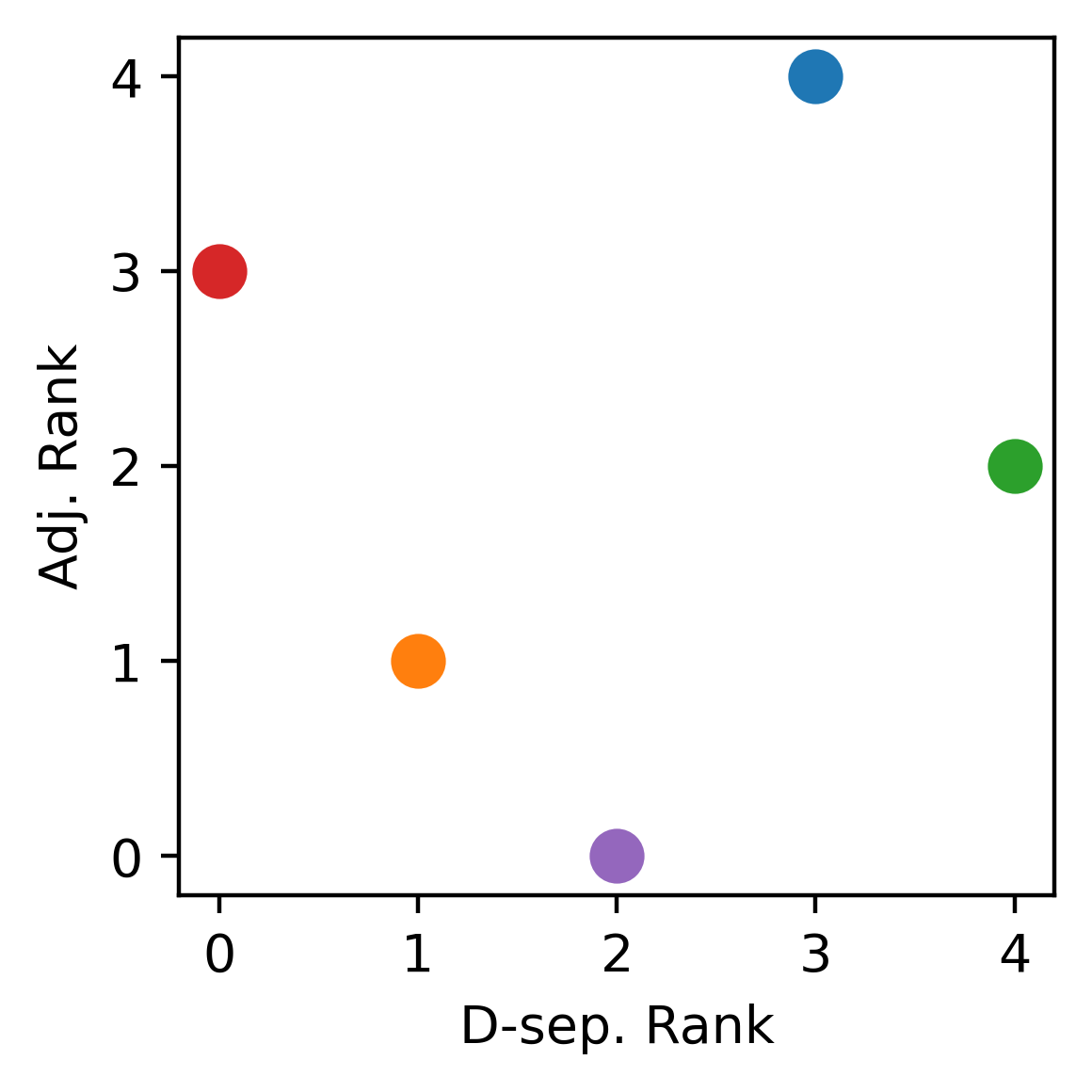}    
    \caption{Causal graph reasoning tasks.}
    \end{subfigure}
    \begin{subfigure}{0.31\textwidth}
    \includegraphics[width=\linewidth]{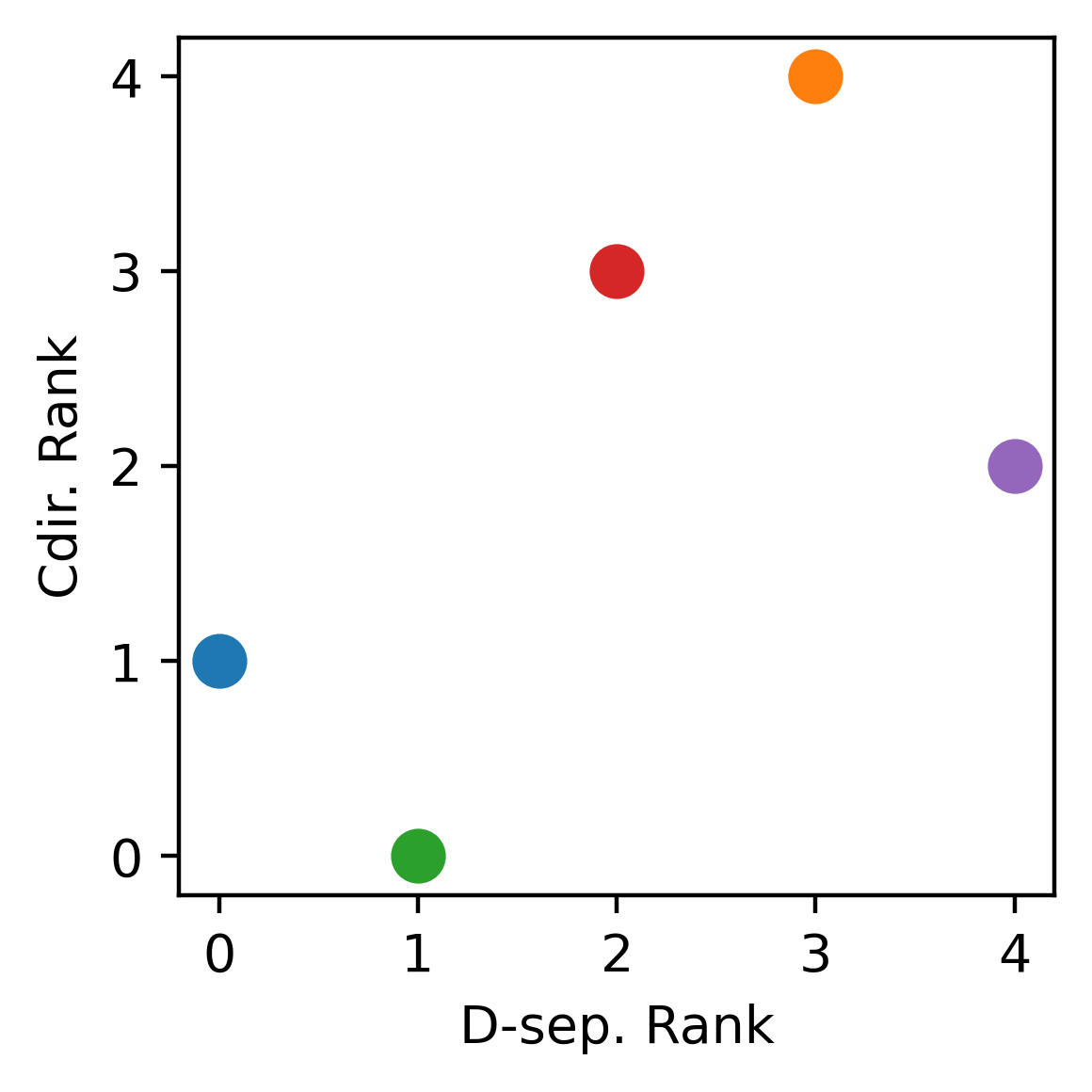}    
    \caption{Causal discovery-based tasks.}
    \end{subfigure}
    \begin{subfigure}{0.31\textwidth}
    \includegraphics[width=\linewidth]{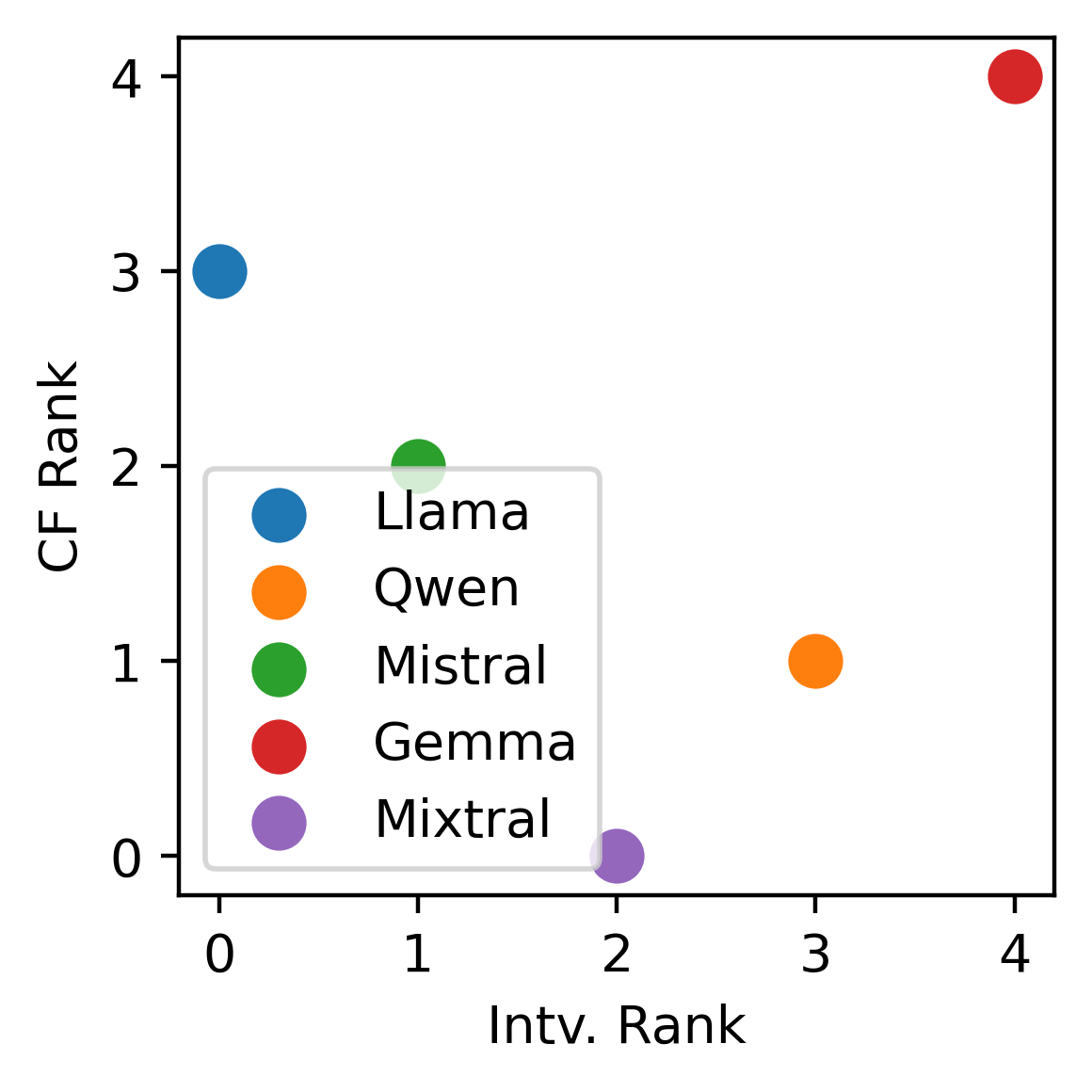}    
    \caption{Causal inference-based tasks.}
    \end{subfigure}
    \caption{Relationships between tasks in the same categories.}
    \label{fig:corr_same_asp}
\end{figure}

\begin{figure}
    \centering
    \begin{subfigure}{0.32\textwidth}
    \includegraphics[width=\linewidth]{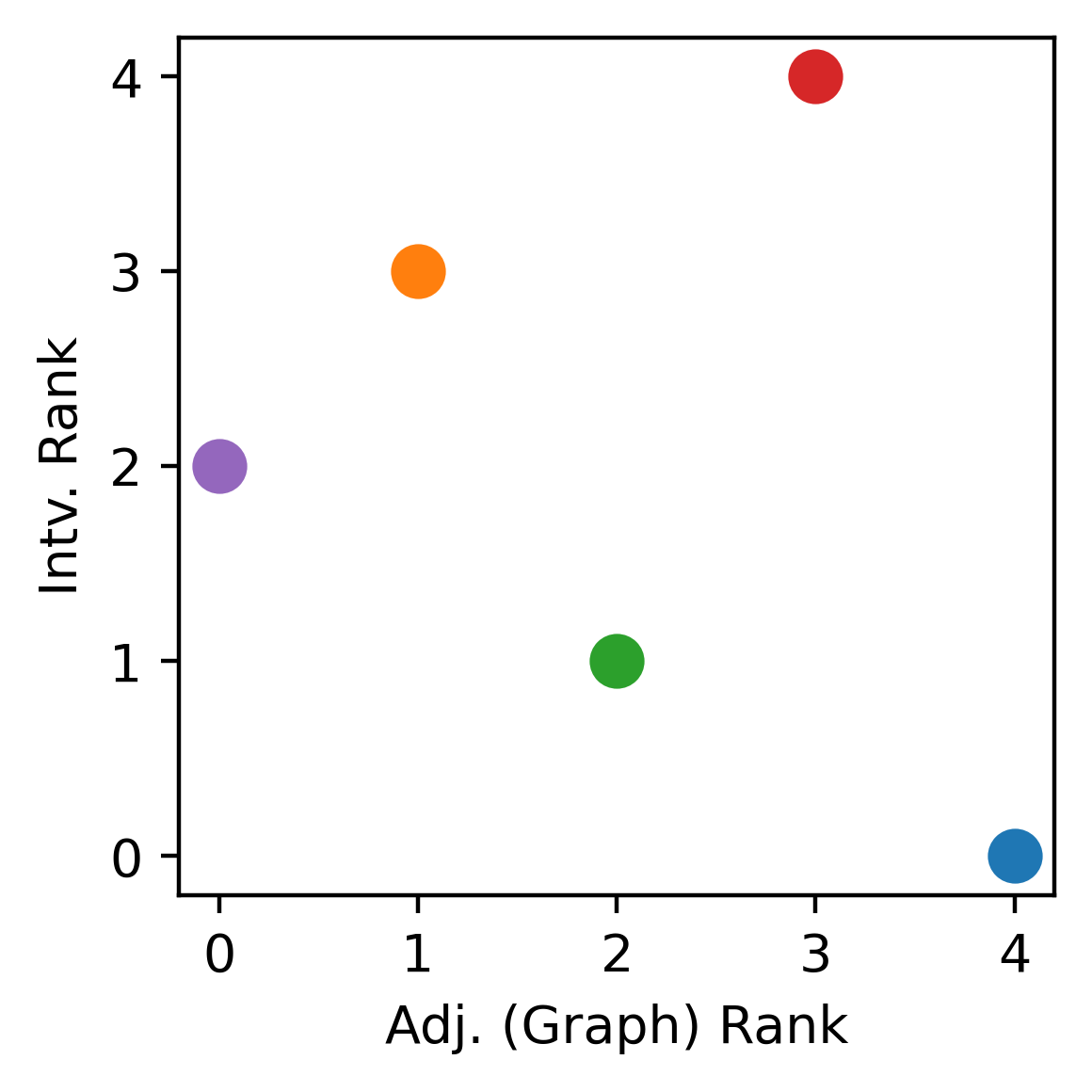}    
    \caption{Graph-based adjacency matrix and interventional distribution estimation.}
    \end{subfigure}
    \hfill
    \begin{subfigure}{0.32\textwidth}
    \includegraphics[width=\linewidth]{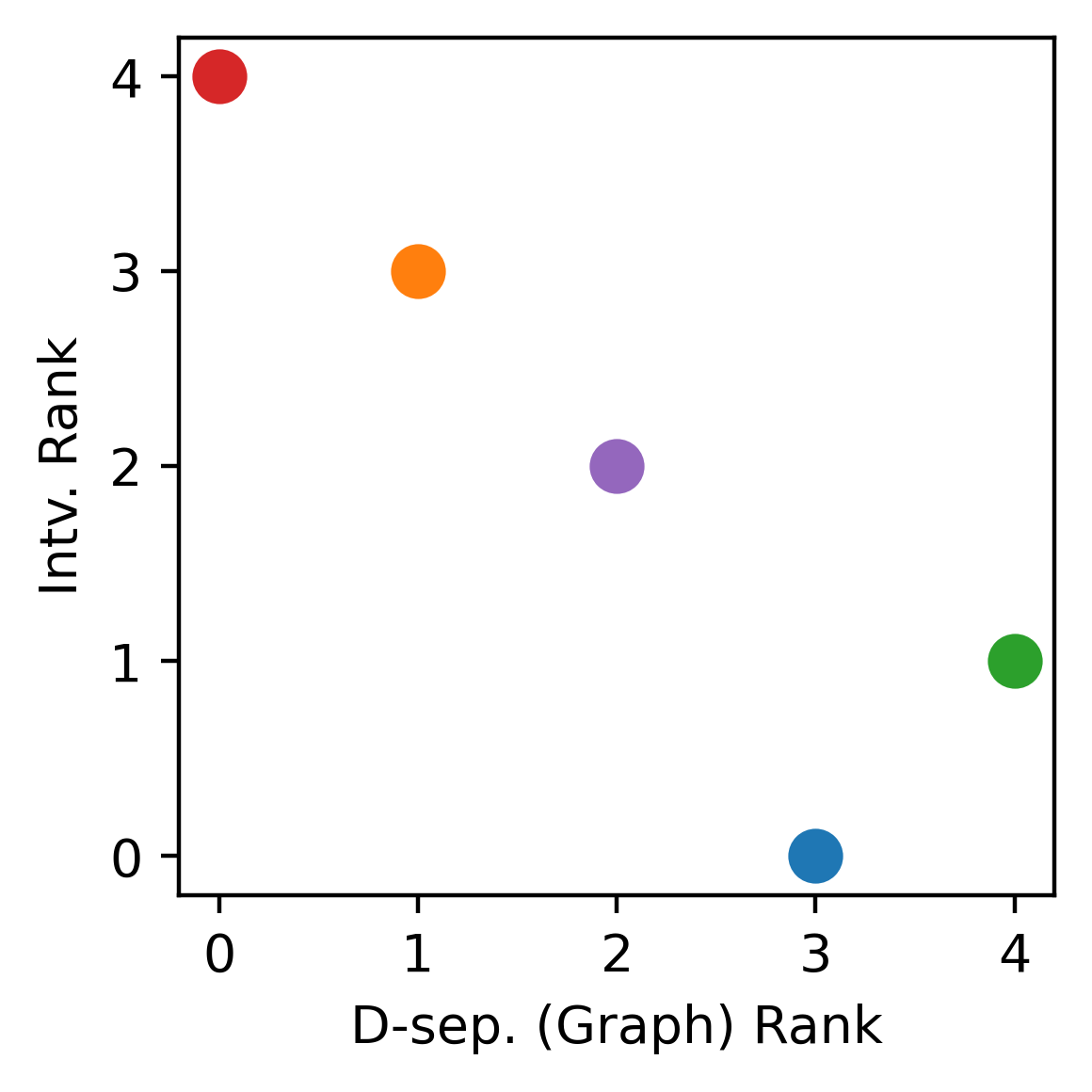}    
    \caption{Graph-based d-separation and interventional distribution estimation.}
    \end{subfigure}
    \hfill
    \begin{subfigure}{0.32\textwidth}
    \includegraphics[width=\linewidth]{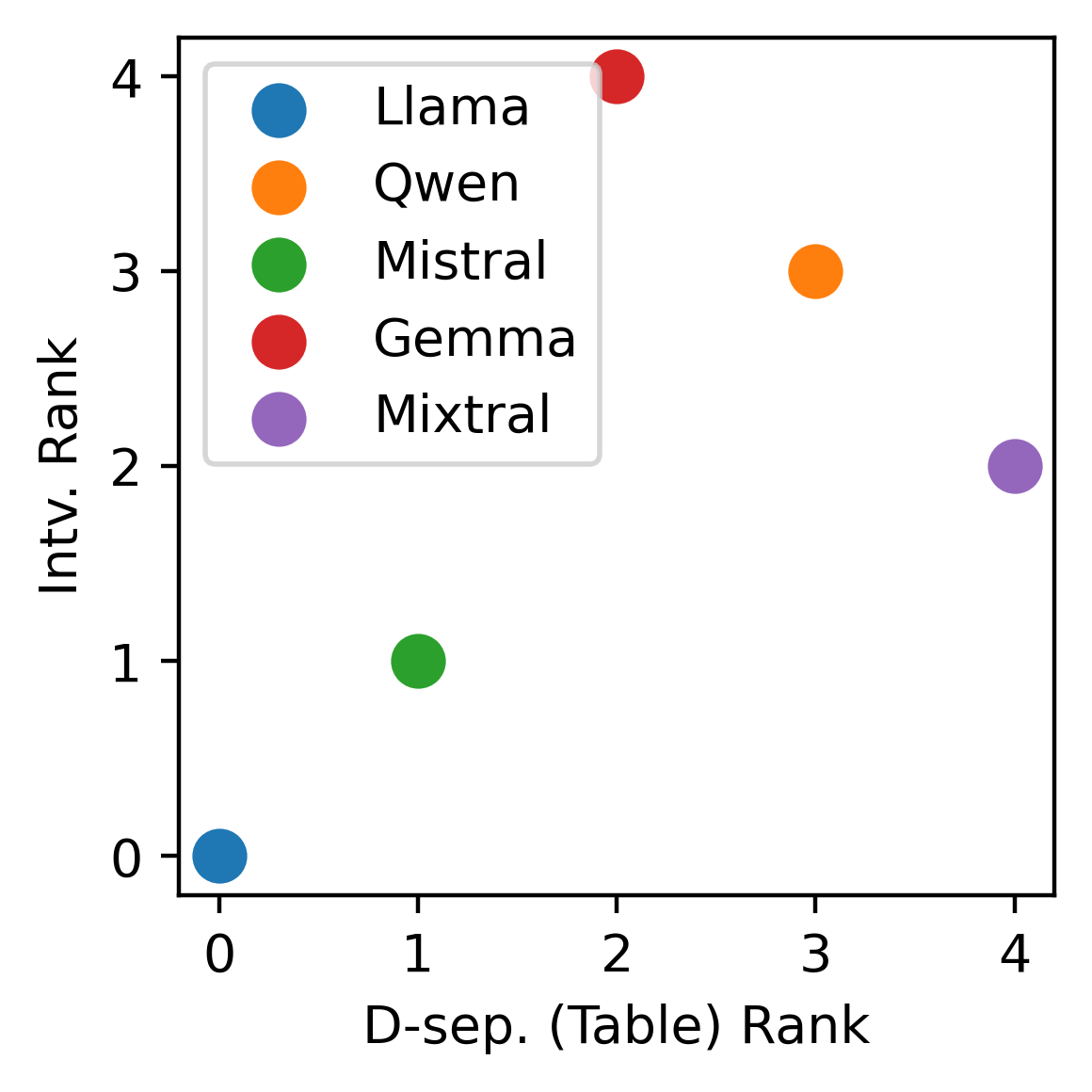}    
    \caption{Table-based d-separation and interventional distribution estimation.}
    \end{subfigure}
    
    \begin{subfigure}{0.32\textwidth}
    \includegraphics[width=\linewidth]{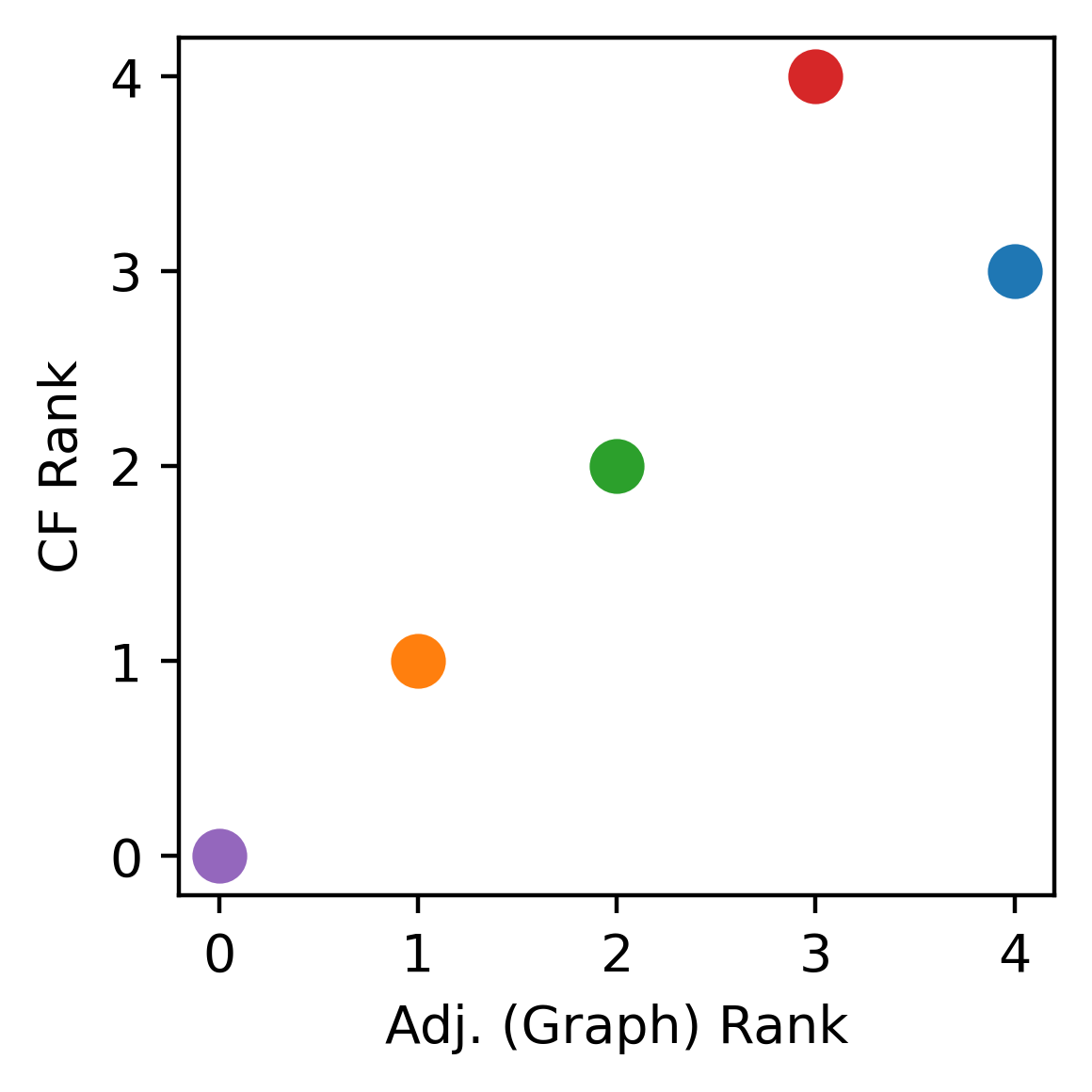}    
    \caption{Graph-based adjacency matrix and counterfactual distribution estimation.}
    \end{subfigure}
    \hfill
    \begin{subfigure}{0.32\textwidth}
    \includegraphics[width=\linewidth]{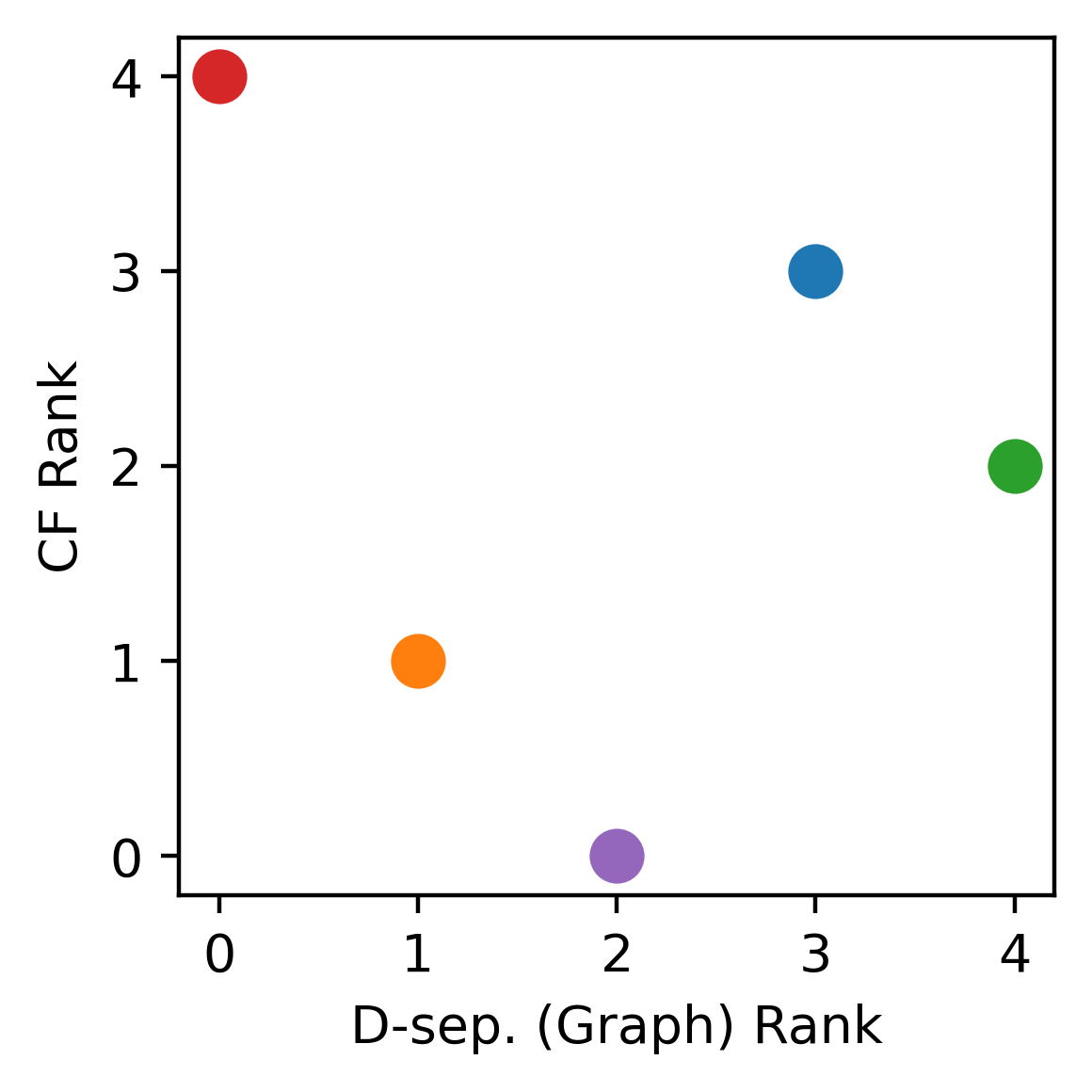}    
    \caption{Graph-based d-separation and counterfactual distribution estimation.}
    \end{subfigure}
    \hfill
    \begin{subfigure}{0.32\textwidth}
    \includegraphics[width=\linewidth]{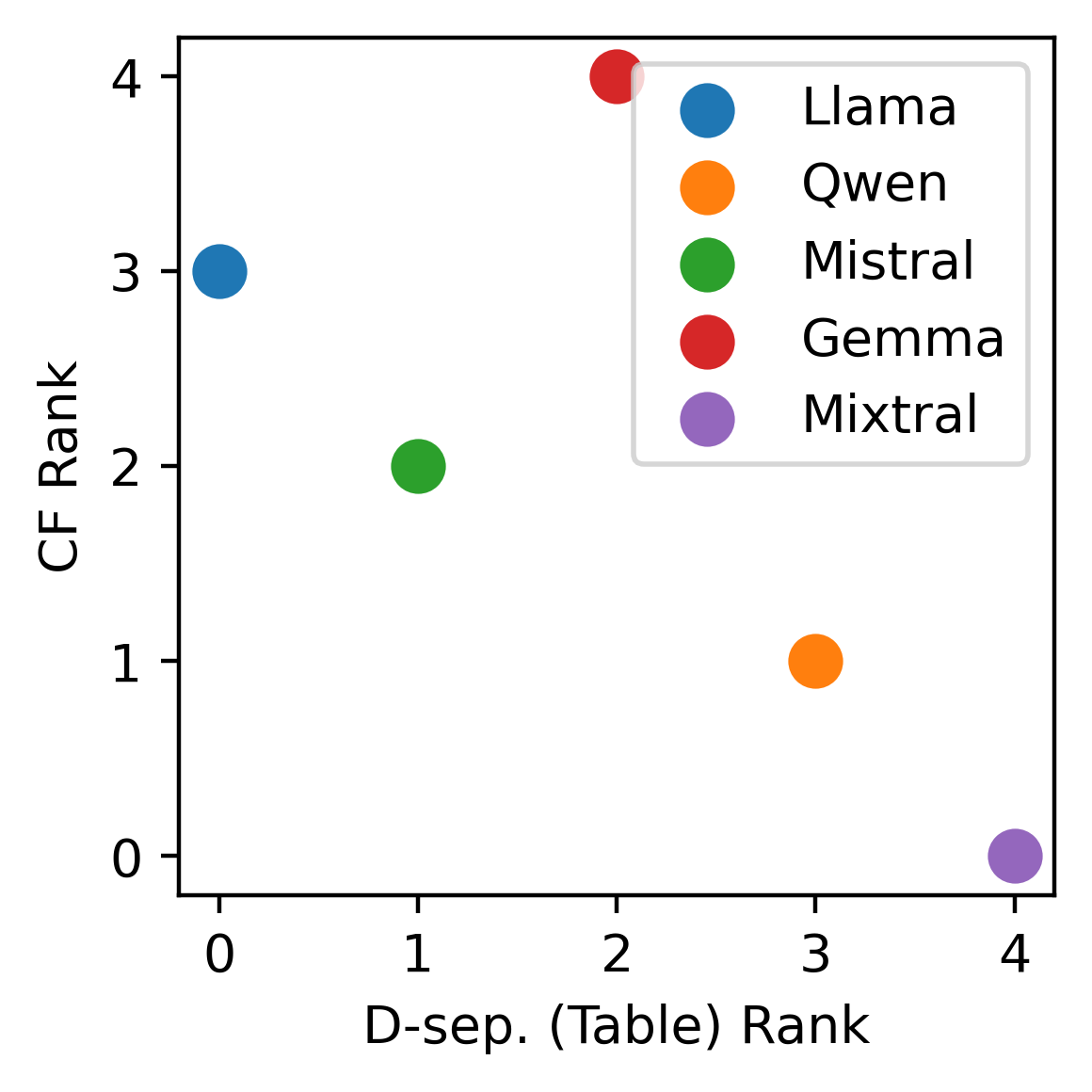}    
    \caption{Table-based d-separation and counterfactual distribution estimation.}
    \end{subfigure}
    \caption{Relationships between causal inference-based and other tasks.}
    \label{fig:corr_ci}
\end{figure}

\begin{figure}
    \centering
    \begin{subfigure}{0.32\textwidth}
    \includegraphics[width=\linewidth]{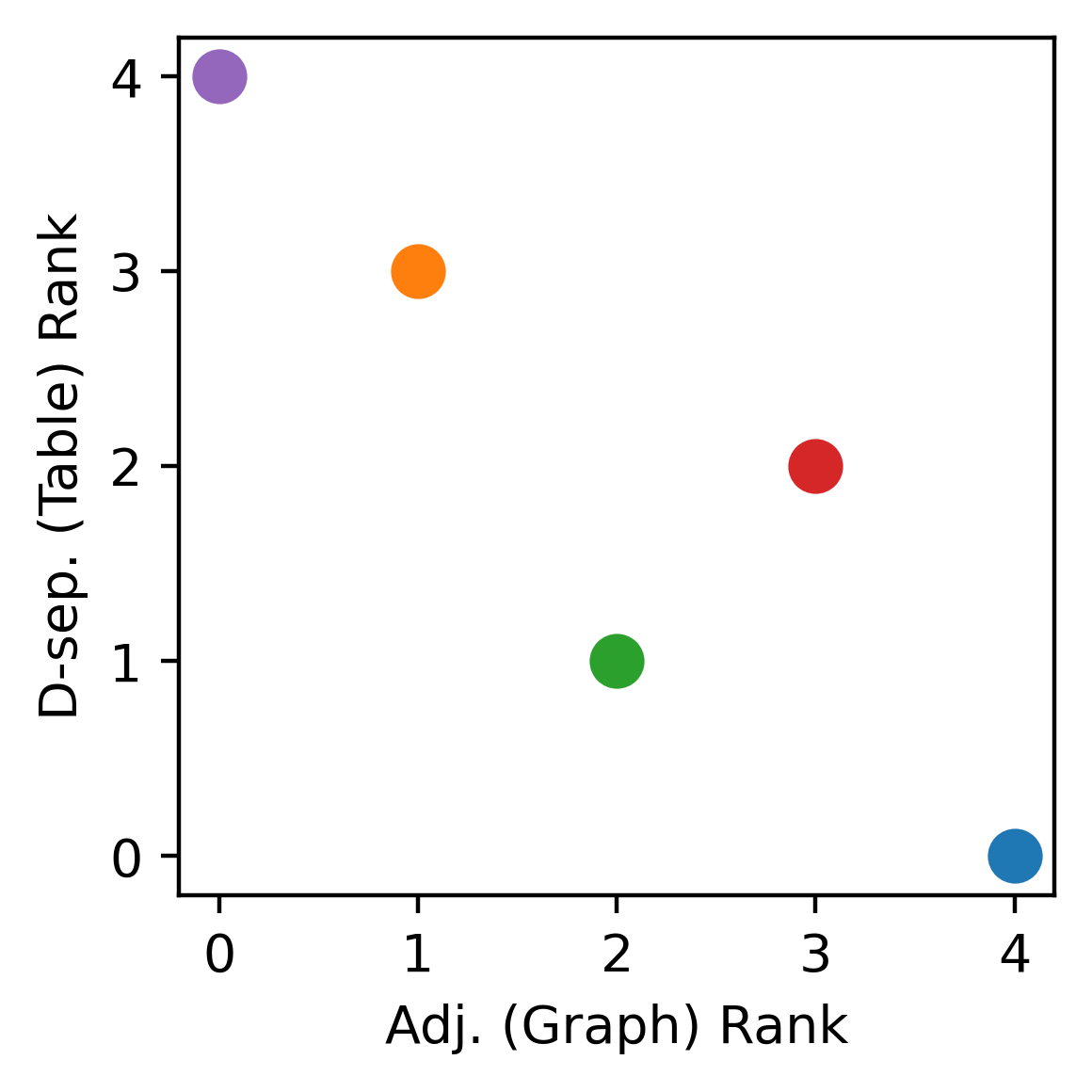}    
    \caption{Graph-based adjacency matrix and table-based d-separation estimation.}
    \end{subfigure}
    \hfill
    \begin{subfigure}{0.32\textwidth}
    \includegraphics[width=\linewidth]{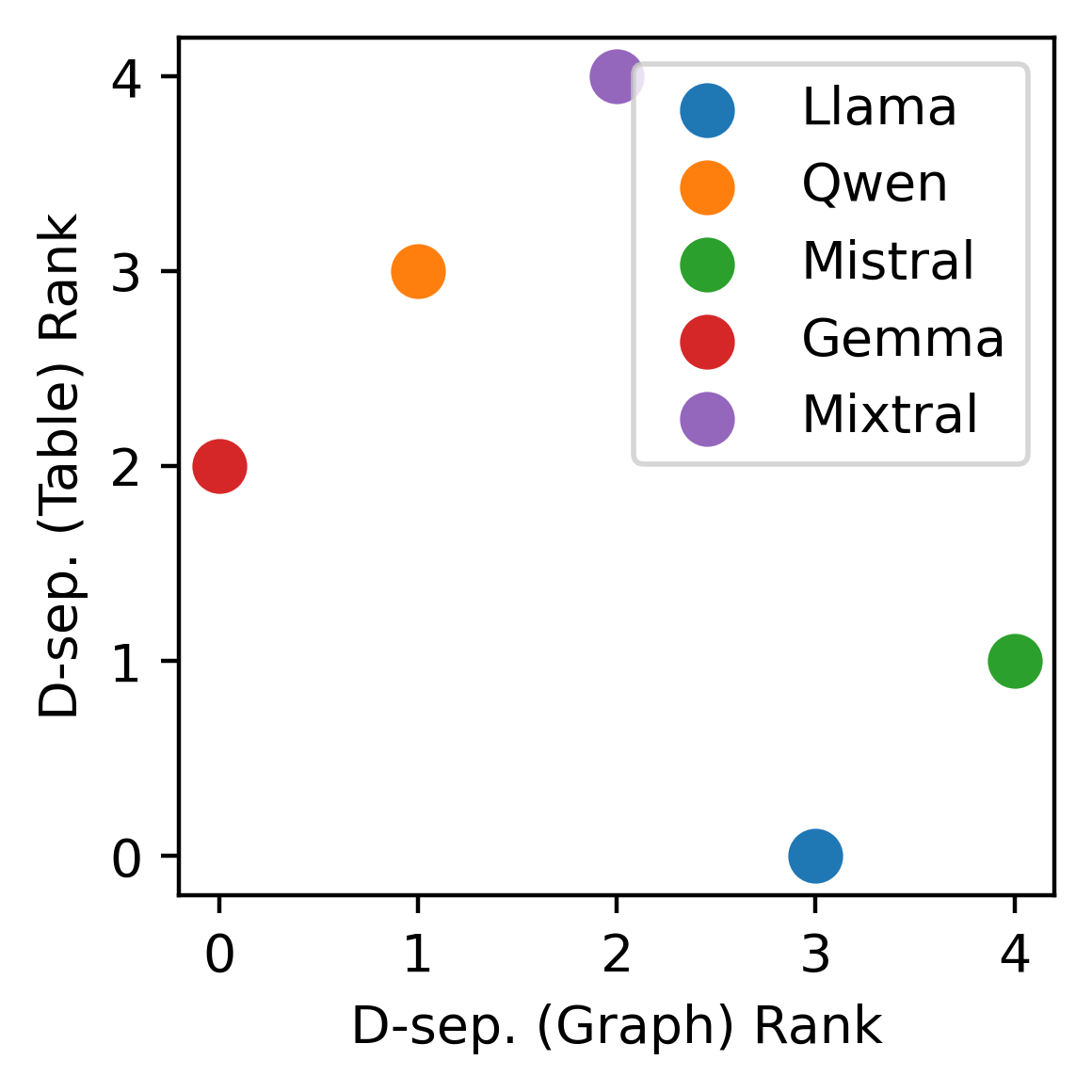}    
    \caption{Graph-based d-separation and table-based d-separation estimation.}
    \end{subfigure}
    \caption{Correlation between causal graph reasoning and causal discovery-based tasks.}
    \label{fig:corr_desp}
\end{figure}

In this section, we investigate how different tasks are related to each other. 
To represent the similarity between tasks, we compute correlation coefficients between ranking vectors with the same order of LLMs. For example, given the order, \llama, \qwen, \mistral, \gemma, and \mixtral, the ranking vector on the d-separation estimation based on causal graphs is $(1, 3, 0, 4, 2)$, where \mistral and \gemma are the best and worst LLMs on this task. Similarly, the ranking vector on adjacency matrix estimation based on causal graphs is $(0, 3, 2, 1, 4)$. By computing the correlation coefficient of these two ranking vectors, one can get the result $0.1$ in Tab. \ref{tab:corr}. A positive correlation suggests that if an LLM performs well on one task, it is likely to perform well on the other task as well; a negative correlation indicates that strong performance on one task is associated with weaker performance on the other; a small or zero correlation implies that performance on one task is not helpful to predict the performance on the other. Moreover, we also visualize the relationships of ranking results of different tasks.
Fig. \ref{fig:corr_same_asp} shows the relationships between tasks in the same categories;
Fig. \ref{fig:corr_ci} shows the relationships between causal inference-based tasks and the other tasks;
Fig. \ref{fig:corr_desp} shows the relationships between causal graph reasoning tasks and causal discovery-based tasks.
We find that the tasks in the same categories have weak correlations. D-separation estimation based on tables has negative correlations with causal graph reasoning tasks. Moreover, intervention distribution estimation has negative relationships with the other tasks while counterfactual reasoning has positive correlation with the other tasks.
On the other hand, there are $5$ LLMs included in the experiments for computing the correlations, which can be more convincing by including more LLMs.

\begin{table}
    \centering
    \caption{Correlation coefficients between the performance of LLMs on tasks. Value $0$ represents no correlation, value $1.0$ represents strong positive correlation, and value $-1.0$ represents strong negative correlation.}\label{tab:corr}
    \begin{tabular}{c|cc|c|cc}
        \hline
         & Adj. (Graph)  & D-sep. (Graph) & D-sep. (Table) & Intv. & CF \\ 
        \hline
        Adj. (Graph) & $-$ & $0.1$ & $\mathbf{-0.9}$ & $-0.3$ & $\mathbf{0.9}$ \\ 
        D-sep. (Graph) & $-$ & $-$ & $-0.5$ & $\mathbf{-0.9}$ & $0.2$ \\ \hline
        D-sep. (Table) & $-$ & $-$ & $-$ & $\mathbf{-0.6}$ & $\mathbf{0.7}$ \\ \hline
        Intv. & $-$ & $-$ & $-$ & $-$ & $0.1$ \\ 
        CF & $-$ & $-$ & $-$ & $-$ & $-$ \\ 
        \hline
    \end{tabular}
\end{table}

\section{Conclusion}
In this work, we propose a benchmark for measuring the capabilities of large language models (LLMs) in causal reasoning. The benchmark evaluates LLMs from multiple perspectives, including causal graph reasoning, knowledge discovery, and decision making. It leverages key concepts of causal discovery and inference to define diverse tasks, such as adjacency matrix, d-separation, causal direction, intervention distribution, and counterfactual distribution. The benchmark tasks are based on serialized tabular data and/or text descriptions of causal graph data. They can reflect the ability of processing complex data and the causal reasoning skills for solving real-world problems where historical data, like patient health measures or scientific observations, are available. To facilitate a comprehensive evaluation, we developed various prompts to assess the open-source LLMs and compare their causal reasoning capabilities. Furthermore, we investigated the relationships between different tasks by analyzing the performance of LLMs across these tasks. Understanding the performance on such tasks and their relationships is crucial for translating benchmark performance into practical real-world scenarios, where the interdependence of tasks reflects the multi-step reasoning and decision-making processes often encountered in real-world problems. 

Due to computational limitations, we were unable to evaluate state-of-the-art large models that require extensive computational resources, despite the comprehensive potential of our benchmark enabled by its synthetic nature. We hope our benchmark will be adopted in both industry and academia to assess the capabilities of state-of-the-art LLMs. Moreover, while our current work focuses on graph and tabular data, which can be easily represented in text format, we recognize the importance of causal reasoning in high-modality contexts. And we plan to advance our benchmark and evaluations to include high-modality settings, such as images. By doing so, we aim to provide a more thorough assessment of LLMs' causal reasoning capabilities to real-world tasks across diverse data modalities.
\section*{Acknowledgement}
The benchmarking results were enabled by resources provided by the National Academic Infrastructure for Supercomputing in Sweden (NAISS), partially funded by the Swedish Research Council through grant agreement no. 2022-06725.

\bibliography{iclr2024_conference}
\bibliographystyle{iclr2024_conference}
\newpage
\end{document}